\documentclass[letterpaper]{article} 
\usepackage{aaai25}  
\usepackage{times}  
\usepackage{helvet}  
\usepackage{courier}  
\usepackage[hyphens]{url}  
\usepackage{graphicx} 
\usepackage{natbib}  
\usepackage{caption} 
\usepackage{algorithm}
\usepackage{algorithmic}
\usepackage{amsfonts}

\usepackage{newfloat}
\usepackage{listings}
\DeclareCaptionStyle{ruled}{labelfont=normalfont,labelsep=colon,strut=off} 
\floatstyle{ruled}
\newfloat{listing}{tb}{lst}{}
\floatname{listing}{Listing}
\usepackage{multirow}
\usepackage{subfigure}
\usepackage[dvipsnames]{xcolor}
\usepackage{amsmath}
\usepackage{subcaption}

\newcommand{\mytitle}{Friends-MMC: A Dataset for Multi-modal Multi-party Conversation Understanding}
\title{\mytitle}
\author{
    Yueqian Wang\textsuperscript{\rm 1},
    Xiaojun Meng\textsuperscript{\rm 2},
    Yuxuan Wang\textsuperscript{\rm 3},
    Jianxin Liang\textsuperscript{\rm 1},
    Qun Liu\textsuperscript{\rm 2},
    Dongyan Zhao\textsuperscript{\rm 1,4 \thanks{Corresponding Author}}
}
\affiliations{
    \textsuperscript{\rm 1}Wangxuan Institute of Computer Technology, Peking University \\
    \textsuperscript{\rm 2}Huawei Noah’s Ark Lab \\
    \textsuperscript{\rm 3}Beijing Institute for General Artificial Intelligence \\
    \textsuperscript{\rm 4}National Key Laboratory of General Artificial Intelligence \\

    \{wangyueqian,liangjx,zhaodongyan\}@pku.edu.cn, \{xiaojun.meng, qun.liu\}@huawei.com, wangyuxuan1@bigai.ai
}

\usepackage{bibentry}

\begin{document}

\maketitle

\begin{abstract}
Multi-modal multi-party conversation (MMC) is a less studied yet important topic of research due to that it well fits real-world scenarios and thus potentially has more widely-used applications. Compared with the traditional multi-modal conversations, MMC requires stronger character-centered understanding abilities as there are many interlocutors appearing in both the visual and textual context. To facilitate the study of this problem, we present Friends-MMC in this paper, an MMC dataset that contains 24,000+ unique utterances paired with video context. To explore the character-centered understanding of the dialogue, we also annotate the speaker of each utterance, the names and bounding bboxes of faces that appear in the video.
Based on this Friends-MMC dataset, we further study two fundamental MMC tasks: conversation speaker identification and conversation response prediction, both of which have the multi-party nature with the video or image as visual context.
For conversation speaker identification, we demonstrate the inefficiencies of existing methods such as pre-trained models, and propose a simple yet effective baseline method that leverages an optimization solver to utilize the context of two modalities to achieve better performance.
For conversation response prediction, we fine-tune generative dialogue models on Friend-MMC, and analyze the benefits of speaker information.
The code and dataset is publicly available and thus we call for more attention on modeling speaker information when understanding conversations.
\end{abstract}


\begin{links}
    \link{Datasets}{https://github.com/yellow-binary-tree/Friends-MMC}
\end{links}

\section{Introduction}

\begin{figure*}[t]
    \centering
    \includegraphics[width=0.85\textwidth]{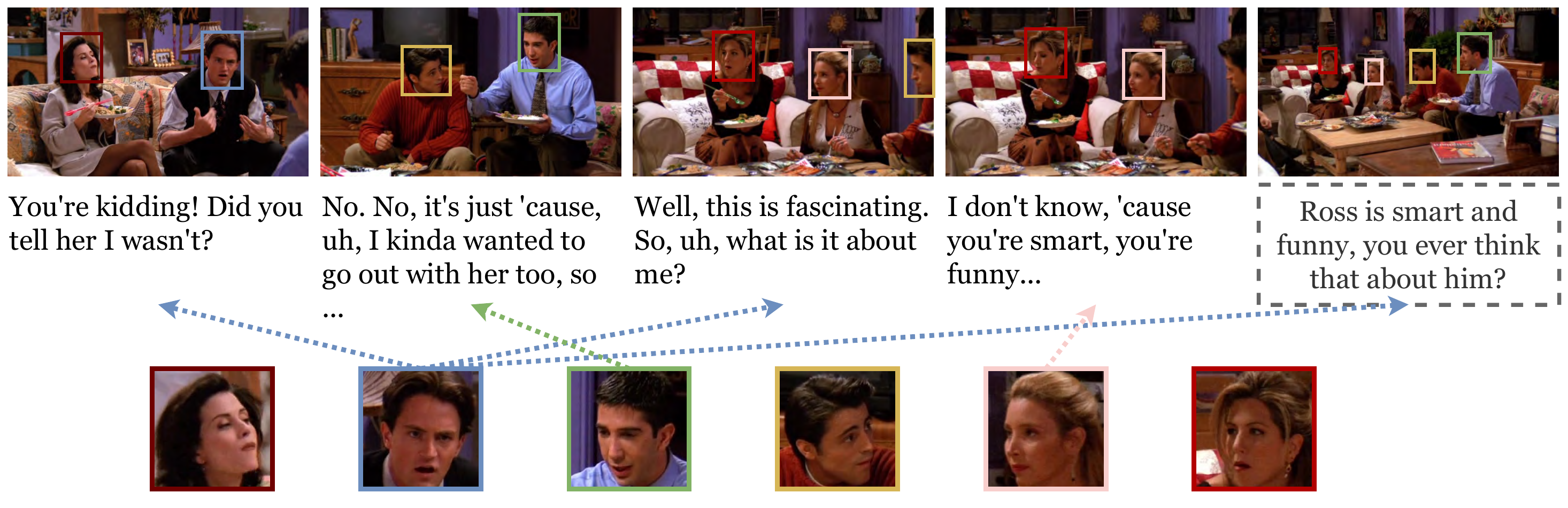}
    \caption{An example of multi-modal multi-party conversation. The task of conversation speaker identification is to infer the dotted arrows pointing from characters to utterances, and the task of conversation response prediction is to infer the last utterance in the dotted rectangular. Only one frame of the video is shown as the visual context to avoid clutter.} \label{fig:task}
\end{figure*}

Multi-modal dialogue systems have attracted extensive attention in recent studies \cite{Zang2021PhotoChat,Zheng2022MMChat,Feng2023MMDialog,Zhu2023MiniGPT4EV,Liu2023VisualIT,Li2023VideoChatCV,Maaz2023VideoChatGPTTD,Li2023MVBenchAC}, especially with the rapid development of multi-modal large language models.
However, there are two main deficiencies of existing works: (1) As most multi-modal datasets are collected from human annotations, LLM generated contents or chat history from social media, these dialogues are mostly presented in question-answer format \cite{AlAmri2019AudioVS} between a human and a system, instead of among several human interlocutors; (2) the speakers are bystanders \cite{Das2016VisualD} and discuss the given visual content such as an image or a video, instead of really being situated in the context.

In contrast, in real-world face-to-face conversations, the interlocutors are situated in the visual \& audio context \footnote{In this work, either images or videos can be used as the visual context. When videos are used, the audios are also included. To avoid cluttering, we keep using ``visual context'' to refer to videos and omit ``\& audio'' in the rest of this paper.}, which means that the conversation and the visual context influence each other rather than interlocutors discussing about a given fixed visual context. Also, there can be more than 2 interlocutors involved, which means that modelling speaker information is crucial for comprehending the conversation.
Understanding face-to-face conversations is an important foundation for achieving embodied artificial intelligence \cite{Zhou2023HowFA,Xu2023ExploringLL,Zhang2023BuildingCEA}. Therefore, we emphasize that multi-modal multi-party conversation is a very important field for real industrial applications yet a less studied topic.

To overcome the shortcomings of existing work on multi-modal conversation, in this paper we extend its scopes and propose a new field of research: multi-modal multi-party conversation (MMC). Compared with traditional multi-modal conversations, MMC differs in the following aspects: (1) The conversation is not only between a user and an assistant, but among an arbitrary number of interlocutors, therefore the information of the speaker should be explicitly modeled; (2) Instead of chatting on images / videos as bystanders, the interlocutors are situated within the visual context, or in other words, the visual context provides rich information about the interlocutors, such as their identities, expressions and actions. 

To foster the study of MMC, we build Friends-MMC, a multi-modal multi-party conversation dataset collected from the famous TV series \textit{Friends}. \footnote{We have investigated the copyright issue of releasing our dataset, it can be released and used for non-commercial purposes. In fact, previous work \cite{Poria2018MELDAM} also created a dataset collected from TV Series \textit{Friends}, and the authors have already released raw videos in their homepage.} An overview of Friends-MMC is shown in Figure \ref{fig:task}. A session in Friends-MMC consists of several turns, each paired with a video clip as visual context, in which the face tracks are detected and classified by character`. Compared to the existing multi-modal dialogue and multi-party conversation datasets, our proposed Friends-MMC have some traits worth emphasizing:


\textbf{a) Modalities} of available data are multiple, including: textual information of each utterance\footnote{In this paper we use the term ``turn'' to denote all contexts of a single turn of speaking, including utterance (\textit{i.e.}, textual content), visual \& audio context, face information and speaker information. We use the term ``utterance'' to denote the textual content only.}, visual \& audio context in forms of frames and videos, face bounding boxes and face character names. Utilizing all of these modalities can be challenging for existing models;
\textbf{b) Conversations} are taken from daily life such as TV series, which are more natural and diverse compared to existing multi-party conversation datasets \cite{Ouchi2016AddresseeAR,Hu2019GSNAG} that are mostly collected from chats on the topic of computers.
\textbf{c) Reasoning} can be very complex. For example, in terms of conversation speaker identification, the speaker may not appear in its corresponding video or frame. Therefore, the preceding or succeeding textual and visual context, as well as their temporal relations, should be taken into account, which is quite difficult to solve for even humans in our experiments.

We focus on two fundamental MMC tasks: \textbf{Conversation Speaker Identification} and \textbf{Conversation Response Prediction}.
The goal of conversation speaker identification is to link the speaker of every turn to the faces in the corresponding visual context. It requires a system to not only infer the speaker of each turn from the textual content, but also understand the visual context in which the dialogue happens. Existing works \cite{Gu2021MPCBERTAP,Meng2018TowardsNSM} of speaker identification mostly focus on text-only multi-party conversation, where only the speaker of the last one or few turns should be predicted given the speaker label of previous turns.
Our paper introduces the multi-modal information of the interlocutors to make this task more challenging and better conform to real-world scenarios where all speaker labels are not available in the dialogue history. 
To accomplish this task, we present our task-specific baseline method including three modality-specific modules which is further introduced later in the ``Conversation Speaker Identification'' Section.
With our task-specific designs, this system can achieve competitive accuracy on conversation speaker identification, far exceeding recent multi-modal pre-trained models, while takes a small amount of computation and a enjoys a high degree of flexibility.

The other task is conversation response prediction, one of the most popular tasks for dialogue modelling. It requires a system to predict the last utterance with respect to the context. Compared with existing response generation task in multi-modal dialogues, the context of MMC is very heterogeneous in modality.
We fine-tune text-only and multi-modal dialogue models on Friends-MMC with different sources of speaker information, to validate that the speaker information is critical to the response prediction of MMCs. In other words, the ability of correctly identifying the conversation speaker can benefit response prediction, which is an issue that has been ignored in existing multi-modal dialogue researches where two speakers simply take turns to speak to each other.

In summary, our contributions are three-fold: (1) We formally propose multi-modal multi-party conversation, a new and valuable field of research, and study two sub-tasks of it: conversation speaker identification and conversation response prediction; (2) We build and release Friends-MMC, a dataset to facilitate the study of multi-modal multi-party conversation; (3) We design a baseline for conversation speaker identification, validate its performance on Friends-MMC, and analyze the benefits of speaker information on conversation response prediction.

\section{Friends-MMC Dataset}
In this section, we describe the dataset collection and annotation procedure for constructing the Friends-MMC dataset, which covers all the 220 episodes from 10 seasons of the TV show \textit{Friends}. The reasons we use \textit{Friends} are: (1) it is a sitcom series, which has numerous conversations that contain diverse topics of daily life; (2) Though having as many as 220 episodes, it has a relatively small number of main characters, which is convenient for automatic face labelling and data cleaning; and (3) It's easy to get publicly available resources like high-quality subtitles that are often manually revised and paired perfectly with the video by a large group of TV fans, which greatly reduces manual labour during the data construction process as well as guarantees the data quality.

The content and speaker of each turn are extracted from transcripts and subtitles\footnote{https://my-subs.co/showlistsubtitles-610-friends \indent~~~https://fangj.github.io/friends}. Faces and their character names in each frame are detected and labelled automatically for the train set (Season 1, 2, and from 4 to 10), and manually for the test set (Season 3) to ensure its accuracy.

\subsection{Construction Process}

\begin{figure*}[t]
    \centering
    \includegraphics[width=0.95\textwidth]{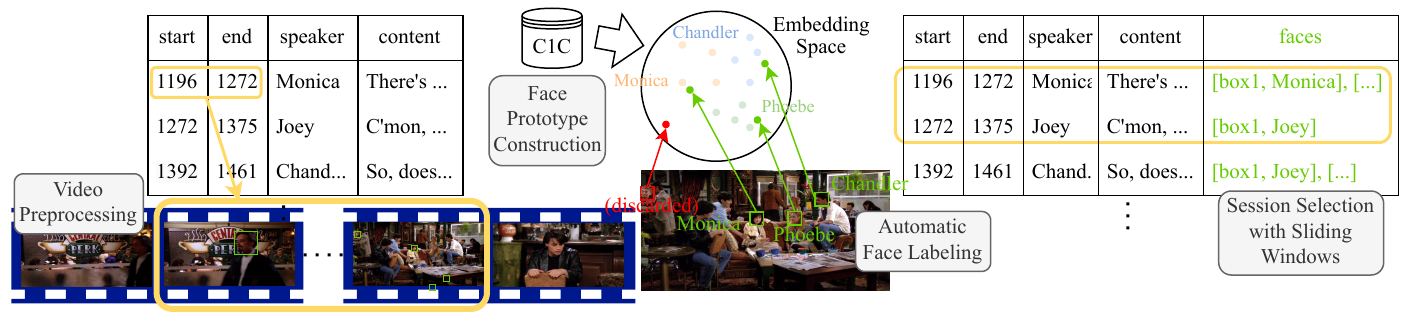}
    \caption{An overview of the construction process of Friends-MMC dataset.} \label{fig:dataset_overview}
\end{figure*}

Figure \ref{fig:dataset_overview} shows the overall construction process of the dataset. Now we introduce every step in details: 

\paragraph{Video Preprocessing.}
We crop a clip from the video according to the start and end timestamp in the subtitle. We use an off-the-shelf face detector \cite{Zhang2017S3FDSS} to detect faces for each frame in the clip. Following \cite{Kalogeiton2020ConstrainedVF}, we merge the faces in adjacent frames into face tracks and thus remove the faces that are not in any track to clean out false positive faces.

\paragraph{Face Prototype Construction.}
C1C \cite{Kalogeiton2020ConstrainedVF} is a dataset with human-labelled face tracks for season 3 of \textit{Friends}. We choose a set of 18 main characters, manually select 20 faces in different viewing angles per character, and encode them using Facenet-512 \cite{Schroff2015FaceNetAU} to get face feature vectors in the embedding space of Facenet-512 as prototypes for each character.

\paragraph{Automatic Face Labelling.}
We automatically label the detected faces tracks with character names by finding their nearest neighbour in the mentioned embedding space. In particular, for each detected face track, we encode each frame in the track with Facenet-512 and calculate the cosine similarity between their feature and all prototypes. If the mean value of the largest 5 cosine similarities is greater than a threshold $t=0.6$ (which is set to maximize the validation accuracy described in the following paragraph), we label this face track with the corresponding character name, otherwise we think this face does not belong to any of the main characters and discard it.

To validate the accuracy of automatic face labelling, we use the same process to detect and label faces for season 3 and compare the results with human-annotated ones from C1C. The verification follows the rule that if the IoU of bounding boxes of an automatically labelled face and a human-annotated face is greater than 0.5, we identify them as a pair of identical faces. Given this threshold, 95\% of all pairs of identical faces are labelled with correct names, which verifies the effectiveness of our automatic face labelling method.

\paragraph{Test \& Test-noisy Set.}
For the test set, we directly use the human-annotated faces in C1C to guarantee the accuracy of face labelling, thus serving as high-quality ground-truths for this test set. Moreover, in order to align with the fact of imperfect face recognition in real-world scenarios and be consistent with the train set, we also create a more challenging test-noisy set by randomly removing 20\% labelled face tracks. 

\paragraph{Image Frame Selection.}
As many recent multi-modal pre-trained models accept images instead of videos as input, we also provide an alternative option of visual context where each turn is paired with only 1 frame. In such case, we select the frame with the most detected faces from each video clip, as well as its face bounding boxes and face names, as the paired visual context of this turn. 

\paragraph{Session Selection with Sliding Windows.}
We use a sliding window of size $m$ to select $m$ adjacent turns as a dialogue session if the following conditions are met: (1) all speakers are from the main characters; (2) the time intervals between all adjacent turns are shorter than 8 seconds, which is a heuristic rule to prevent selecting turns from different scenes. Therefore, we use $m = \{5, 8\}$ to create two types of sessions with different context lengths (5 turns and 8 turns). Note that different dialogue sessions may contain the same turn, as it belongs to different contexts and thus the preceding or succeeding textual and visual contents differ.

\subsection{Dataset Statistics}
Dataset statistics are shown in Table \ref{tab:stat}. Apart from the basic statistics, we also count the proportion of speakers whose faces are not detected in the current clip or frame, or even not appear in all the $m$ clips or frames within this entire session. 

As table \ref{tab:stat} shows, the test-noisy set includes a significantly larger number of speakers not in the current clips (or frames for image as input) than the test set. Therefore, this case is more challenging for speaker identification task, as the candidate model needs to really understand the conversation and find out more clues from the context rather than only the current clips or frames to infer who is the real speaker.

\begin{table*}
    \centering
    \fontsize{9}{9}\selectfont
    \setlength{\tabcolsep}{1mm}
    \begin{tabular}{c|ccc|ccc}
        \hline
         & \multicolumn{3}{|c|}{5 turns} & \multicolumn{3}{|c}{8 turns} \\
         & train & test & test-noisy & train & test & test-noisy \\
        \hline
        \# sessions & 13584 & 2017 & 2017 & 8730 & 1325 & 1325 \\
        \# unique turns & 21092 & 3069 & 3069 & 16990 & 2480 & 2480 \\
        \# words in utterance & 18.87 & 20.28 & 20.28 & 18.71 & 20.42 & 20.42 \\
        \# speakers in each session & 2.83 & 2.85 & 2.85 & 3.43 & 3.47 & 3.47 \\
        \hline
        \# face tracks per clip & 2.41 & 3.12 & 2.50 & 2.39 & 3.14 & 2.52 \\
        avg. secs per face track & 2.31 & 2.71 & 2.72 & 2.30 & 2.74 & 2.73 \\
        \% speakers not in current clip & 13.43 & 1.03 & 19.26 & 13.51 & 1.10 & 18.93 \\
        \% speakers not in all clips & 6.13 & 0.17 & 1.13 & 5.57 & 0.14 & 0.44 \\
        \hline
        \# faces per frame & 1.61 & 2.20 & 1.76 & 1.60 & 2.21 & 1.78 \\
        \% speakers not in current frame & 24.05 & 6.52 & 25.64 & 24.15 & 6.42 & 25.30 \\
        \% speakers not in all frames & 9.53 & 1.01 & 3.32 & 7.45 & 0.42 & 1.37 \\
        \hline
    \end{tabular}
    \caption{Dataset Statistics of Friends-MMC. We provide a train set, a test set and a more challenging test-noisy set.}
    \label{tab:stat}
\end{table*}

\section{Conversation Speaker Identification} \label{sec:csi}
\subsection{Task Introduction}
Conversation speaker identification requires models to identify the speaker of each turn given the textual and visual context. 
Existing works on speaker identification mostly focuses on the text-only multi-party conversation. It often asks models to predict the speaker of the last few turns, given the dialogue history and the speaker of previous turns.
However, in real-world scenarios, speaker labels are usually either available for all turns or not available for any turn.
To simulate it, when we extend the task to multi-modality, we provide the basic visual information of interlocutors so that the speaker of each turn can be predicted.

However, too many decisive clues for identifying the speaker, such as voice characteristics and facial movements, can also be provided by the video and audio. This will cause models to tend to only rely on the rich video information and ignore the dialogue context, which deviates from our original motivation of promoting researches on conversation systems. Therefore, we also propose an alternative setting: providing only one frame and no audio as the visual context. In this setting, the basic information of interlocutors such as identities, expressions, and the scene they are in, can also be provided by this frame, and it leaks less decisive clues and shortcuts for directly identifying the speaker.

\subsection{Baseline Method}
To effectively utilize all the modalities of the proposed benchmark dataset, including visual, audio, textual and face tracks, we propose a baseline method, which consists of three modules: 1) a visual model $M_1$ to recognize speaking faces, 2) a text model $M_2$ to analyse multi-speaker relations based on dialogue contexts, and 3) a quadratic binary optimization problem solver to combine their results and thus identify the speaker of each turn. This modular design makes our system enjoy a high degree of flexibility, as one can easily change the visual model or the text model with alternative ones when different contextual information (e.g., using image or video as visual input) is provided. 
Figure \ref{fig:model} shows the overview of our proposed method, and we introduce each module in the following paragraphs.

\subsubsection{Visual Model}
We use a visual model to predict the probability of each face belonging to the current speaker individually: $p_{face} = M_1(face) \in (0, 1) $.
There are two different visual contexts:
1) If the provided visual context is an image frame, we use an inception model \cite{Szegedy2014Inception} pre-trained on VGGFace2 \cite{Cao2017VGGFace2AD} and fine-tuned on the train set of Friends-MMC as the visual model $M_1$, and $face$ is an image of the face cropped using bounding box $b$. When fine-tuning on Friends-MMC, the speaking label of a face $y_{face}$ is set to $1$ if the character name $c$ of this face is identical to the speaker name $y$, and $0$ otherwise: $y_{face} = \mathbf{1}[c = y]$. We use the cross-entropy classification loss as the training objective.
2) If the provided visual context is a video, we use TalkNet \cite{Tao2021TalkNet}, a state-of-the-art active speaker detection model, as the visual model $M_1$, and $face$ is a video of the face cropped using bounding box sequence $b$, accompanied by the audio of the same period. Note that using TalkNet in a zero-shot manner is good enough, we try to fine-tune it on the train set of Friends-MMC as like the CNN model, but it does not result in better performance.

\subsubsection{Text Model} \label{sec:m_2}
We use a DeBERTa-v3 \cite{he2021debertav3} first trained on Ubuntu Dialogue Corpus \cite{lowe-etal-2015-ubuntu} and then fine-tuned on the train set of Friends-MMC as $M_2$ to predict whether every two utterances in a dialogue session are spoken by the same speaker. The intuitive reason behind it is that for some utterances, it is hard to identify the speaker solely by feeding the video or frame into $M_1$. We thus try to conjecture its speaker by finding whether it likely shares the same speaker with another utterance, for which we have confidences or prior knowledge to infer its speaker. In addition, $M_2$ helps to achieve a relatively global optimum speaker inference by considering all turns together.

Given a dialogue session consists of $m$ utterances, we prepend an \texttt{<bos>} token to each utterance as the input of $M_2$ as like: \newline $\texttt{<bos>}u_1 \cdots \texttt{<bos>}u_m$. We use the last hidden state of each \texttt{<bos>} $h_i$ as the representation of each utterance, and use a head layer to calculate the similarity of every two representations:

\begin{equation*}
p_{sim}^{ij} = \sigma(W_2 \text{GeLU}(W_1[h_i; h_j; |h_i - h_j|] + b_1) + b_2)
\end{equation*}

where $i, j = 1, \cdots, m$, and $(W_1, b_1, W_2, b_2)$ are learnable parameters. $\sigma$ is the sigmoid activation function, and $p_{sim}^{ij} \in (0, 1)$ is a scalar that denotes the probability of two utterances spoken by the same person. The loss function is defined as:
\begin{equation*}
\mathcal{L}_{M_2} = MSE(p_{sim}, y_{sim}) + MSE(p_{sim}, p_{sim}^T)
\end{equation*}

where $y_{sim} \in \{0, 1\}^{m \times m}$ is the ground truth label of whether any two utterances are from the same speaker, and $MSE$ denotes mean squared error loss.

\subsubsection{Quadratic Binary Optimization Problem Solver}
This module aims to integrate the outputs of both visual ($M_1$) and textual ($M_2$) models for speaker identification. In particular, we design two matrices using the outputs of these two models, and address it as the quadratic binary optimization.

For each dialogue session, we first obtain a candidate speaker set by recording all faces appeared in every frame / video: $\mathbf{C} = \{c_1, \cdots, c_{l}\}$. We construct a vision reward matrix $\mathbf{B} \in \mathbb{R}^{l \times m}$ of selecting a character $c_i$ as the speaker of the turn $u_j$. If the face of $c_i$ appears in the frame / video $v_j$, $b_{ij}$ is set to the probability of that face as a speaking face predicted by $M_1$, otherwise $b_{ij} = 0$. Apparently, $\mathbf{B}$ can only express those situations that the speaker appears in the corresponding frame / video. To address those problems, we construct another text reward matrix $\mathbf{A} \in \mathbb{R}^{m \times m}$ to reveal the reward for assigning the same speaker to two turns $u_i$ and $u_j$. We first use $M_2$ described in the previous subsection and get a similarity matrix $p_{sim}$. However, if we simply use this similarity matrix $p_{sim}$ as the reward matrix $\mathbf{A}$, since all elements in $p_{sim}$ are larger than 0, the optimization solver tends to assign the same speaker to every turn to get the maximum rewards. To avoid which, we subtract the similarity matrix with the mean value of its elements as the reward matrix, \textit{i.e.,} $\mathbf{A} = p_{sim} - \text{mean}(p_{sim})$

\begin{figure}
    \centering
    \includegraphics[width=\linewidth]{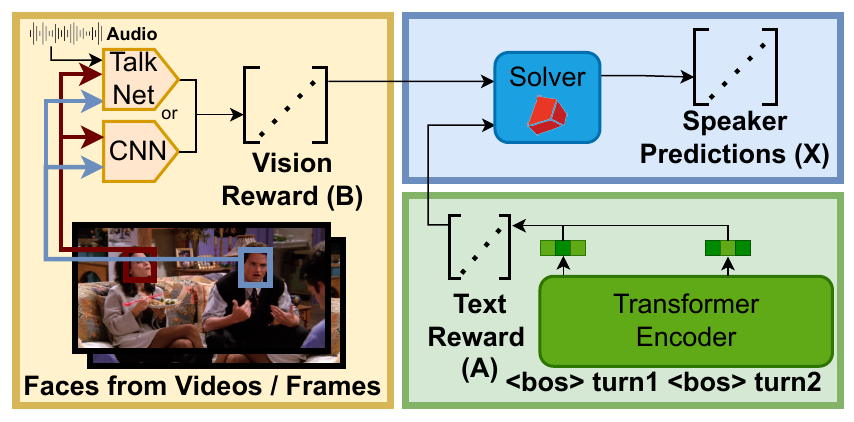}
    \caption{Model overview of the three modules in different colors: the visual ($M_1$) is yellow, the textual ($M_2$) is green, and the optimization solver taking vision and text reward matrix as input is blue.}
    \label{fig:model}
\end{figure}

As Figure \ref{fig:model} shows, with $\mathbf{A}$ and $\mathbf{B}$ in hand, the task of multi-modal multi-party speaker identification can be represented by a quadratic binary optimization problem:
\begin{align*}
\text{Maximize} \quad & f(X) = (1-\alpha)X^TAX + \alpha XB \\
\text{s.t.} \quad & X \in \{0, 1\}^{m \times l}, \\
\quad & \sum_{j=1}^{l} X_{ij} = 1, \quad i = 1, 2, \ldots, m
\end{align*}

where $\alpha$ is a hyperparameter to control the weight of two rewards and is selected according to the performance on a validation set held-out from the train set. We use 0.8 for frame as visual context, 0.7 for video as visual context, and 0.2 when ground truth labels of the text model are provided ($M_2^\dag$). By now, this problem can be easily solved using optimization problem solvers like \cite{gurobi}, which adaptively makes decisions based on the output of $M_1$ and $M_2$. The reason we use an optimization solver instead of an end-to-end pre-trained model is that this task of identifying speakers still remains challenging to use the general attention mechanism of pre-trained models to fuse different modalities. 

\subsection{Experiment Results}
\subsubsection{Implementations}
We conduct experiments in five settings with information of different modalities used: (1) image (frame) only; (2) video only; (3) text only; (4) image and text; and (5) video and text. In image only setting (1), we use the model $M_1$ CNN to predict one face from all detected faces from the frame as the speaker. If there are no faces, we randomly choose a character from the candidate speaker list. The video only (2) is almost the same as (1) but uses $M_1$ TalkNet to predict one face from the video. In the text only setting (3), we compare $M_2$(DeBERTa-v3) with 3-shot GPT 3.5 using its in-context learning ability to perform this task. For $M_2$, although it is good at judging whether two utterances are said by the same speaker, it is not trained to identify the speaker for a single utterance. Therefore, it can only make guesses according to the relations between sentences. GPT 3.5, however, possesses some ability to understand the candidate speaker list and identifying names from utterances, so it can make full use of the textual context to achieve more accurate reasoning.
In image-text setting (4), the jointed $M_1$ (CNN) + $M_2$ model is used together with a quadratic binary optimization solver, and we also try to replace the output of $M_2$ with ground truth labels (denoted by $M_2^\dag$) to explore bottlenecks and possible improvement directions.  We also test GPT4-o \cite{openai2024gpt4o} under this setting.
Video-text setting (5) is almost the same as (4), but again CNN is replaced by TalkNet for $M_1$.

We also fine-tune three popular and powerful multi-modal pre-trained models: Violet \cite{Fu2021VIOLETE}, LLaVA \cite{Liu2023VisualIT} and Emu \cite{Sun2023Emu}. As LLaVA only accepts one image at a time during pre-training, we also use one single turn (one utterance and one frame) during fine-tuning and testing, neglecting information from adjacent turns. Violet and Emu, on the other hand, are pre-trained with videos or image-text interleaved data, so we can directly use all 5 or 8 turns as input.

To check how much useful information can be provided by using frame as the visual context, we also report the human performance of this task. we randomly sample 80 dialogue sessions from each (5 turns / 8 turns) test set, provide dialogue contents, frames, face bounding boxes \& names to participants, and ask them to select a speaker for each turn from the candidate speaker set (\textit{i.e.,} the characters that appear in all frames).
This process requires intensive efforts from humans, according to their post-interview, as the task of selecting speakers requires careful observation and a thorough understanding of the dialogue contents. We thus only perform the human studies on the test set, since we believe the human performance on the test-noisy set should be apparently worse. See appendix for details of all the baselines mentioned above.

\subsubsection{Main Results}

\begin{table} \
    \centering
    \fontsize{9}{9}\selectfont
    \setlength{\tabcolsep}{1mm}
    \begin{tabular}{cc|cc|cc}
    \hline
    & & \multicolumn{2}{|c|}{5 turns} & \multicolumn{2}{|c}{8 turns} \\
    & & & noisy & & noisy \\
    \hline
    \multirow{2}{*}{0} & random & 31.82 & 32.61 & 28.54 & 29.03 \\
     & (std.dev.) & (0.25) & (0.47) & (0.49) & (0.27) \\
    \hline
    \multicolumn{5}{l}{\textit{Frame Only}} \\
    \hline
    1 & $M_1$(CNN) & 72.88 & 63.72 & 72.90 & 62.51 \\
    \hline
    \multicolumn{5}{l}{\textit{Video Only}} \\
    \hline
    2 & $M_1$(TalkNet) & 80.89 & 70.91 & 81.00 & 70.50 \\
    \hline
    \multicolumn{5}{l}{\textit{Text Only}} \\
    \hline
    3 & $M_2$ & 33.24 & 33.85 & 29.09 & 29.33 \\
    4 & GPT 3.5 (3-shot) & 37.21 & 37.24 & 33.35 & 32.81 \\
    \hline
    \multicolumn{5}{l}{\textit{Use image and text modality}} \\
    \hline
    5 & Violet & 32.66 & 33.16 & 27.73 & 28.86 \\
    6 & LLaVA v1.5-13B  & 46.30 & 42.39 & 45.73 & 41.41 \\
    7 & Emu-14B & 61.76 & 58.23 & 60.96 & 56.46 \\
    8 & $M_1$(CNN) + $M_2$ & 75.81 & 68.61 & 74.53 & 67.21 \\
    9 & $M_1$(CNN) + $M_2^\dag$ & 84.90 & 78.01 & 90.80 & 83.93 \\
    10 & GPT-4o (0-shot) & 66.36 & 65.60 & 63.64 & 61.02 \\
    11 & Human & 82.25 & - & 84.49 & - \\
    \hline
    \multicolumn{5}{l}{\textit{Use video and text modality}} \\
    \hline
    12 & $M_1$(TalkNet) + $M_2$ & 83.21 & 74.12 & 83.60 & 75.00 \\
    13 & $M_1$(TalkNet) + $M_2^\dag$ & 90.88 & 83.09 & 95.10 & 89.69 \\
    \hline
    \end{tabular}
    \caption{Accuracy on the test and test-noisy set of Friends-MMC. $M_1$ and $M_2$ denote the visual and textual model of our baseline method, respectively. For $M_1$, we use CNN or TalkNet to take image or video as input. \dag ~indicates that we use ground truths instead of textual model outputs ($M_2$) to serve as upper bounds.} \label{tab:main_results}
\end{table}


Results can be found in Table \ref{tab:main_results}: (1) visual context acquired by the vision model $M_1$, including which face appears in the frame and looks like a speaking face, serves as the most critical clues, shown by the performance of $M_1$ (line 1, 2). We can conclude that this speaker identification task is still vision dominant, especially when videos are provided. 
(2) Speaker relations acquired by the text model $M_2$ also play a vital supporting role to bring a significant improvement of $2\%\sim5\%$ from $M_1$ to $M_1$ + $M_2$ (line 8 \textit{vs.} line 1, line 12 \textit{vs.} line 2). The textual context benefits this task not only by providing dialogue contents, but also aims for more real scenarios where the speaker does not appear in the scene. What's more, the benefit is greater when the paired visual model is more accurate. It makes sense that one has to accurately identify speakers of some turns before it is able to identify other turns using this speaker relation information. 
(3) Directly fine-tuning a multi-modal pre-trained model (line 5, 6, 7) or using proprietary multi-modal models (line 10) is not a good option despite we have tried different models with various parameter sizes. We believe the key aspects that are essential to solve this problem remain difficult to be understood by these candidate models, as this task differs a lot from the original training objectives. It also may be due to the reason that this speaker identification task is difficult to be formatted as the proper and shorten input of the model and thus to be easily learned.
(4) Comparing the line 8-9 and 12-13, our current method of fine-tuning $M_2$ model to predict whether two utterances are spoken by the same speaker still has a lot of room for improvement. This strong upper bound result also evidences that our designed objective for the textual module is meaningful.

The influence of the hyper-parameter $alpha$ to the results is listed in the appendix.


\section{Conversation Response Prediction}
\subsection{Task Introduction}
Conversation response prediction aims to predict the textual content of the last turn given the visual context of all turns and the textual content of all previous turns.
In this study, we demonstrate that speaker information is critical to the response prediction task of multi-modal multi-party conversations. 
Though predicting the next utterance in multi-modal dialogues has been a widely studied topic, none of the previous work focuses on the specific nature of multi-party dialogues, \textit{i.e.,} taking the speaker information into account.

We hypothesize that the potential advantages come from speaker information in two aspects: (1) The model may learn the speaking style of each speaker. When the speaker information like name is available, it can be used as a \textbf{global speaking style indicator} for generating better responses; (2) Speaker information can be a \textbf{local context indicator} to infer the response is from which person's point of view, \textit{i.e.,} should be consistent with which utterance in the local conversation.

To validate these two aspects, we constructed 3 test sets: (a) a test set with random speaker names: speakers of all utterances are randomly assigned. Here the potential advantages of (1) and (2) are both broken. 
(b) a test set with random history speaker names: similar to (a) but we keep the speaker name of the utterance we need to predict unchanged. Here only the potential advantage of (2) is broken.  (c) a test set with shuffled names: we replace all names in both speaker information and utterance content with another name according to a predefined random shuffle mapping, \textit{e.g.,} replace ``Ross'' with ``Joey'', and ``Joey'' with ``Chandler''. Here only the potential advantage of (1) is broken. Note that in all cases the speaker information of the train set is not modified. 

Our experiments are in the following settings: train and test without speaker name; train with ground truth speaker information and test with ground truth or random or shuffled or our identified speaker name ($M_1$ + $M_2$). Note that when using our $M_2$ model, the last utterance to predict is removed from the input, \textit{i.e.,} the speaker of top $m-1$ turns is predicted using both visual and textual context, and the speaker of the last turn is predicted using visual context.
In addition, we finetune Emu-14B as a dialogue model with visual input, and Llama2-7B as a dialogue model without visual input.
As mentioned in \cite{Wang2023LearnTW}, it is often hard to predict the next utterance solely by the dialogue history, hence the text generation metrics (\textit{e.g.,} rouge) may not be very valid when evaluating on our movie dialogue dataset. Therefore, following \cite{Wang2023LearnTW}, we build a response selection task to evaluate models instead of using generative metrics. We randomly select 9 other utterances from the test set as negative responses, and ask the model to select the one with the lowest perplexity from totally 10 (plus ground truth) candidates as output. 

\subsection{Experiment Results}

\begin{table}[t]
    \centering
    \fontsize{9}{9}\selectfont
    \setlength{\tabcolsep}{1mm}
    \begin{tabular}{cc|c|c}
        \hline
        Model & Speaker & 5 turns & 8 turns \\
        \hline
        \multirow{7}{*}{Llama2-7B} & No & 30.69 & 36.98 \\ 
         & Random & 31.23 & 43.32 \\ 
         & Random History & 31.63 & 43.40 \\ 
         & Shuffled & 35.20 & 48.60 \\ 
         & Ground truth & 36.89 & 49.36 \\ 
         & $M_1$(CNN) + $M_2$ & 34.16 & 45.81 \\ 
         & $M_1$(TalkNet) + $M_2$ & 34.56 & 46.64 \\ 
        \hline
        \multirow{7}{*}{Emu-14B} & No & 30.49 & 31.09 \\ 
         & Random & 29.35 & 31.55 \\ 
         & Random History & 29.45 & 31.25 \\ 
         & Shuffled & 33.02 & 35.17 \\ 
         & Ground truth & 34.06 & 36.30 \\ 
         & $M_1$(CNN) + $M_2$ & 31.98 & 33.89 \\ 
         & $M_1$(TalkNet) + $M_2$ & 32.97 & 34.64 \\ 
        \hline
    \end{tabular}
    \caption{Accuracy of conversation response prediction by selecting one from a set of ten utterances as candidates. } \label{tab:response}
    \label{tab:stat}
\end{table}

As shown in Table \ref{tab:response}, regardless of which pre-trained model is used, or the length of dialogue context, in all cases adding speaker information always improves the performance. Comparing to using the random speakers and random history speakers, the performance with shuffled speakers is much closer to the one with ground truth speakers. Therefore, in terms of our hypothesis on the two potential advantages that come from speaker information,
this result indicates that speaker information mainly works by indicating the local context instead of serving as a global speaking style indicator. Overall, our baseline method of speaker identification also benefits the response prediction task as it achieves a significant better performance than the ones with no or random speakers.

\section{Related Works}
\subsection{Multi-party Conversations}
Multi-party conversations (MPC), as opposed to two-party conversations, is a more practical and challenging scenario which studies conversations that involve more than two interlocutors. Research on MPC often includes three sub-topics: speaker prediction, utterance prediction, and addressee prediction. 
\cite{Ouchi2016AddresseeAR,lowe-etal-2015-ubuntu} built MPC datasets from Ubuntu technical dialogues and proposed baseline models for MPC tasks. Recent studies on MPC usually train and evaluate models jointly on those three objectives. \cite{Gu2021MPCBERTAP} propose MPC-BERT, which fine-tunes BERT \cite{Devlin2019BERTPO} on several self-supervised tasks, and achieve state-of-the-art results on the above MPC tasks. \cite{Su2022SpeakerC} identify speakers of utterances by clustering them with pairwise relations encoded by a dialogue content encoder. GIFT \cite{Gu2023GIFTGF} revises the model structure of transformer encoders to make the self-attention layer be aware of the information flow of MPC. 
Details regarding MPC can be found in these recent surveys \cite{Gu2022WhoSW,Ganesh2023ASO}.

\subsection{Multi-Modal Dialogue Datasets}
There have been a number of works on building multi-modal dialogue datasets \cite{Das2016VisualD,Mostafazadeh2017ImageGroundedCM,Shuster2020ImageChatEG,AlAmri2019AudioVS,Zheng2022MMChat,Zang2021PhotoChat,Feng2023MMDialog}.
Despite the diversity in modality (image or video) and the position of the visual information in the dialogue history, all above datasets are limited as the interlocutors are outside the visual contexts rather than ``situated'' inside them, and only includes two interlocutors instead of being ``multi-party''.

Dialogue in movie/TV series is a typical data source of multi-party conversation with ``situated'' visual context. Recent large-scale movie dialogue datasets include OpenViDial \cite{Meng2020OpenViDialAL,Wang2021OpenViDial2A} and VSTAR \cite{Wang2023VSTAR}. However, these datasets do not have any kind of speaker information, which hinders a deeper-level understanding of the conversation. Perhaps the dataset most similar to ours is MELD \cite{Poria2018MELDAM}, which is also a speaker-aware multi-modal multi-party dialogue dataset collected from \textit{Friends} but focuses only on emotion recognition, and does not annotate faces in the visual context.

\section{Conclusion}
We work on a new field of research, multi-modal multi-party conversation (MMC), to bridge the gap between existing researches on multi-modal dialogue and real-world applications such as face-to-face conversations or meetings. We build Friends-MMC, an MMC dataset in which each utterance is paired with video context, speaker, face bounding box and face name annotation. We propose two new tasks of MMC, namely conversation speaker identification and conversation response prediction, and design a baseline to solve both tasks. Our future directions include collecting more diverse data rather than movies, and exploring ways to make better use of speaker information for response generation.

\section{Acknowledgments}
The authors thank the kind suggestions and support from AI Data Technology Laboratory of Huawei Noah's Ark Lab.

\bibliography{aaai25}

\clearpage
\appendix
\section*{Appendix}
\bigskip

\section{Analysis of Reward Weights $\alpha$ in Conversation Speaker Identification}
\begin{figure}[t]
    \centering
    \subfigure[Test set performance on 5 turns dataset.]{
    \includegraphics[width=0.45\columnwidth]{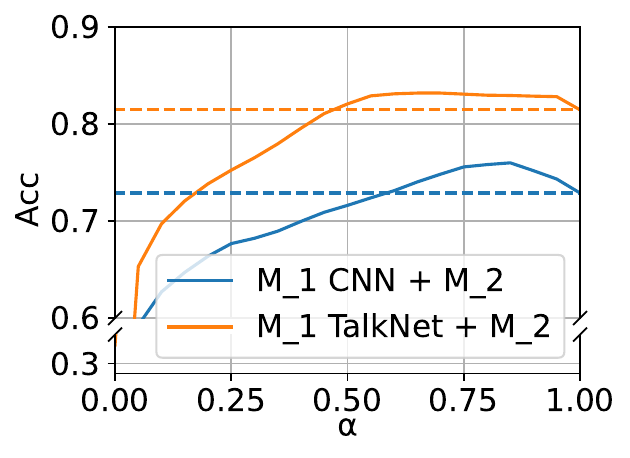}}
    \subfigure[Test set performance on 8 turns dataset.]{
    \includegraphics[width=0.45\columnwidth]{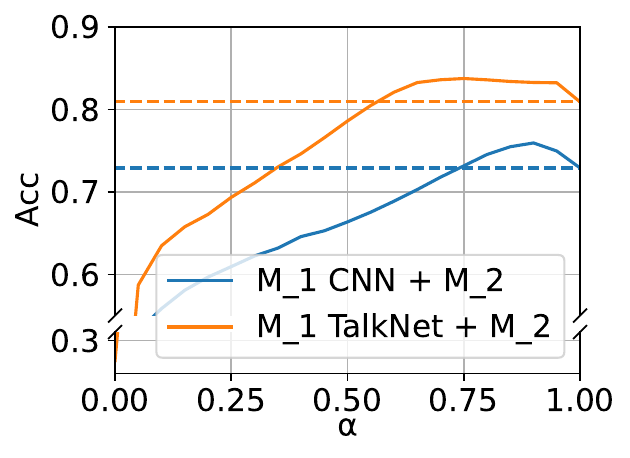}}
    \vspace{-1em}
    \caption{The change of accuracy with respect to $\alpha$. The dotted horizontal line shows the performance of only using the visual model.} \label{fig:alpha}
    \vspace{-1em}
\end{figure}

Figure \ref{fig:alpha} shows the change of accuracy with respect to $\alpha$, a key hyper-parameter which controls the weight of the scores provided by the visual model $M_1$ and the text model $M_2$. Note that when $\alpha = 0$, the task reduces to using only the text model $M_2$, and when $\alpha = 1$, the task reduces to using only the visual model $M_1$. In a considerable range of $\alpha$ values, introducing the results of $M_2$ improves the overall accuracy, compared with using $M_1$ only. It verifies that textual contexts certainly contribute to this speaker identification task.

\section{Details of Baseline Methods of Conversation Speaker Identification}
\subsection{GPT 3.5}
We use in-context learning to perform 3-shot inference with GPT 3.5. \textcolor{orange}{Instruction}, \textcolor{blue}{input} and \textcolor{green}{expected target} we use is as follow:

\bigskip \hrule \smallskip
\textcolor{orange}{You are listening to a conversation among a group of people. You will be provided with a name list and the content of conversation, and need to guess which people in the name list speaks each turn of the conversation. Answer one name for each turn in the dialogue, \texttt{[num turns]} comma-seperated names in all.} \\
\textcolor{blue}{Name list: \texttt{[candidate 1]}, \texttt{[candidate 2]}, \dots \\ Conversation (one utterance per line):  \\ \texttt{[utterance1]} \\ \texttt{[utterance2]} \\ \dots \\ Answer: \texttt{[speaker 1]}, \texttt{[speaker 2]}, \dots \\ \\ \texttt{[several more examples]} \\ \\ Name list: \texttt{[candidate 1]}, \texttt{[candidate 2]}, \dots \\ Conversation (one utterance per line):  \\ \texttt{[utterance1]} \\ \texttt{[utterance2]} \\ \dots} \\ \textcolor{blue}{Answer:} \textcolor{green}{\texttt{[speaker 1]}, \texttt{[speaker 2]}, \dots}
\smallskip \hrule \bigskip

If GPT 3.5 generates more than \texttt{[num turns]} names, we only keep the first \texttt{[num turns]} names as its predictions. If GPT 3.5 generates less than \texttt{[num turns]} names, or generates names not in the candidate list, we pad its prediction / replace the name not in the candidates list with names randomly selected from the candidate list.

\subsection{GPT-4o}
We evaluate the performance of GPT-4o in zero-shot manner. The prompt we use is as follow:

\bigskip \hrule \smallskip
\textcolor{orange}{You are listening to a conversation among a group of people. Their names are: ``\textcolor{blue}{\texttt{[candidate 1]}, \texttt{[candidate 2]}, \dots}'', which is labeled on the frames. The visual context of the conversation is also provided in \textcolor{blue}{\texttt{[num\_turns]}} images, one image per turn. You need to guess which people in the name list said each sentence. Answer one name for each turn in the dialogue, for example, "Alice, Bob, Alice, Carol". \\ The conversation is (one turn per line):} \\
\textcolor{blue}{\texttt{[utterance1]} \\ \texttt{[utterance2]} \\ \dots} \\ \textcolor{blue}{Answer:} \textcolor{green}{\texttt{[speaker 1]}, \texttt{[speaker 2]}, \dots}
\smallskip \hrule \bigskip

\subsection{Violet}
We fine-tune all the parameters of Violet. The input format is:

\bigskip \hrule \smallskip
\texttt{[patches of all images] [candidate 1] [candidate 2] ... [CLS] [utterance 1] [CLS] [utterance 2] ...}
\smallskip \hrule \bigskip

and calculate the cosine similarity between the representation of each utterance (\textit{i.e.,} the last hidden state of the \texttt{[CLS]} before it) and each speaker (\textit{i.e.,} the last hidden state of the candidate speaker name \texttt{[candidate x]}).

\subsection{Emu}
We use LoRA \cite{hu2022lora} as an paramter-efficient method to fine-tune Emu-14B.
As Emu was pre-trained with interleaving image-text data, we also use interleaving visual context and text context as its input. The input format is as follow:

\bigskip \hrule \smallskip
\textcolor{orange}{Choose the name of the speaker for the each of the \texttt{[num turns]} turns given the dialog history, frames, and faces in the frames.} \\ \textcolor{blue}{Image of turn 1: \texttt{[image 1]} \\ Content of turn 1: \texttt{[utterance 1]} \\ Faces in turn 1: \texttt{[bbox 1] [name 1]}, \texttt{[bbox 2] [name 2]}, ... \\ Image of turn 2: \\ ...} \\ \textcolor{orange}{Speakers of the above \texttt{[num turns]} frames:} \\ \textcolor{green}{\texttt{[speaker 1]}, \texttt{[speaker 2]}, ...}
\smallskip \hrule \bigskip

\subsection{LLaVA}
We use LoRA \cite{hu2022lora} as an paramter-efficient method to fine-tune LLaVA-v1.5-13B. As LLaVA is pre-trained with inputs with a single frame at a time, we also use only 1 turn at a time as its input. The input format is as follow:

\bigskip \hrule \smallskip
\textcolor{blue}{\texttt{[image]}} \\ \textcolor{orange}{Choose the name of the speaker for this sentence:} \textcolor{blue}{\texttt{[utterance]}} \\ \textcolor{orange}{Given the names and bounding boxes of the faces for the people that appear in the visual context where the dialog takes place:} \textcolor{blue}{\texttt{[bbox 1] [face 1]}, \texttt{[bbox 2] [face 2]}, ...} \textcolor{green}{\texttt{[speaker]}}
\smallskip \hrule \bigskip

\subsection{Human Performance}
Here we provide instructions given to participants for the human performance experiment:

\bigskip \hrule \smallskip
The two folders each contain 20 pieces of data, with 5 or 8 rounds of conversation

Each piece of data contains 5 (or 8) screenshots of TV series. In each screenshot, the green word indicates the content of the line, and the red word indicates the name of the person appearing in the screenshot
There is a line of blue characters at the bottom, which is a collection of all the red characters above. Your task is to select the person from the blue name for each of the 5 (8) lines that is most likely to be the speaker of the line.
Note that the speaker may not appear in the corresponding screenshot, so you may need to make a comprehensive judgment based on the context information. If you feel that you can't judge accurately, guess an answer, guessing is also an important ability of humans.

There are also two files, 5\_turns-pred.txt and 8\_turns-pred.txt in the package. The two files now have a row each, representing the results of my annotation of 19 .jpg this piece of data.
You need to add 20 lines to each of these two files in the same format to indicate the results of your annotation of 60.jpg-79 .jpg these 20 pieces of data. That is, first enter a number to indicate which data you are marking, and then enter 5 (8) personal names to represent speakers for each line, separated by spaces.
If you find it too cumbersome to enter the entire person's name, you can just enter the first two letters. For example, "ro" for "ross", "ch" for "chandler", etc.
\smallskip \hrule \bigskip

\end{document}